\pdfoutput=1

\documentclass[runningheads]{llncs}

\usepackage[linesnumbered]{algorithm2e} 
\usepackage{todonotes}
\usepackage{comment}

\usepackage{cite}
\usepackage{hyperref}

\usepackage{graphicx}

\begin{document}
\title{
Hybrid Ensemble-Based  Travel Mode Prediction
}
\author{Paweł Golik\orcidID{0009-0003-1254-6879} \and
Maciej Grzenda\orcidID{0000-0002-5440-4954} 
\and 
Elżbieta Sienkiewicz\orcidID{0000-0001-9468-4674}}
\authorrunning{Golik et al.}
\institute{Warsaw University of Technology, Faculty of Mathematics and Information Science, Koszykowa 75, 00-662 Warszawa, Poland \\
\email{\{Pawel.Golik.stud, Elzbieta.Sienkiewicz, Maciej.Grzenda\}@pw.edu.pl}
}

\maketitle              %
\begin{abstract}

Travel mode choice (TMC) prediction, which can be formulated as a classification task, helps in understanding what makes citizens choose different modes of transport for individual trips. This is also a major step towards fostering sustainable transportation. As behaviour may evolve over time, we also face the question of detecting  concept drift in the data. This necessitates using appropriate methods to address potential concept drift. In particular, it is necessary to decide whether batch or stream mining methods should be used to develop periodically updated TMC models. %

To address the challenge of the development of TMC models, we propose the novel Incremental Ensemble of Batch and Stream Models (IEBSM) method aimed at adapting travel mode choice classifiers to concept drift possibly occurring in the data. It relies on the combination of drift detectors with batch learning and stream mining models. We compare it against  batch and incremental learners, including methods relying on active drift detection. Experiments with varied travel mode data sets representing both city and country levels show that the IEBSM method both detects drift in travel mode data and successfully adapts the models to evolving travel mode choice data. The method has a higher rank than batch and stream learners. 

\end{abstract}

\section{Introduction}
The growth in the volume of data streams has caused 
data mining methods, which analyze bounded and stationary datasets, to be potentially unable to %
adapt to shifting data patterns and dynamic phenomena~\cite{moa}.
In addition to the regular fluctuations and random variations in the data, the 
\textit{concept drift} phenomenon~\cite{gama2010knowledge, moa}  is frequently observed due to reasons such as seasonality~\cite{surveyieee}. 
A fast and potentially infinite \textit{data stream} can be viewed as the output of a stochastic process that produces data based on a specific probability distribution at a given time~\cite{surveyieee}. This distribution may change over time. More precisely, concept drift arises when the distribution $p(x)$ and/or $p(y|x)$ changes between consecutive labelled stream instances $\{x,y\}$~\cite{surveyieee}. 

The travel mode choice datasets typically include features such as duration and reason of the trip, traveller's attributes, e.g., age and gender, represented in vector $\mathbf{x}$. Each trip record includes as well a travel mode selected by the traveller ~\cite{hillel2021, hagenauer2017, garcia2022}, i.e., a class $y_i\in Y$, which includes using a car, public transport, cycling and walking. As trips are made over time, it is not obvious whether classifiers predicting travel mode should be trained with batch methods assuming stationary settings i.e., fixed $p(y|x)$ and $p(x)$ or stream mining methods addressing concept drift \cite{gomes2017adaptive}.
Stream mining algorithms offer the advantage of continuous incremental model training. 
This 
enables a model to adapt to concept drift,
but it fundamentally alters the training process and the employed machine learning algorithms by including either passive or active adaptation to concept drift ~\cite{moa}.
Alternatively, models can be trained on batches of data, a typical approach. In this scenario, models can effectively detect patterns present in batches due to the ability to iterate over data multiple times, often yielding improved outcomes. Still, their effectiveness might be reduced when the incoming data shifts~\cite{
gandomani2020role}. 

Given the formulation of a TMC problem as a classification task, \cite{hagenauer2017, hillel2021},
a question is posed whether online classifiers should be applied to learn possibly evolving models reflecting the evolving decisions of travellers, or whether the magnitude of concept drift(s) is not sufficient to use online learners rather than batch models. The answer is likely to depend on how stationary the process is in different cities or countries and  may even change with time. Hence, we propose an ensemble method combining multiple batch and online methods to reduce the risk of selecting an under-performing method.
Furthermore, the method we propose, can utilize multiple batch-learning models, bringing an extra advantage. It enables the use of distinct configurations of drift detection and batch model retraining strategies, referred to as {\em drift handling strategies}, for each batch learner. 

Hence, in this work, we propose the novel Incremental Ensemble of Batch and Stream Models (IEBSM) method
for predicting the preferred mode of transport. We evaluate both the method and baseline learners using various real datasets of successively recorded trips provided by respondents. %
As the distribution $p(x^i)$ of some features present in the TMC data sets is likely to change with time, trip data offer a compelling illustration of unending and changing data streams. 
The primary contributions of this work are as follows:
\begin{itemize}
    \item We propose the novel IEBSM ensemble method combining drift detectors with batch and online learners. The method automates the use of multiple batch and online methods, drift detection and the retraining of batch models. The experiments we performed show that the IEBSM method yields performance gains over batch and online methods for various travel mode choice tasks. It provided highest ranked TMC models. We provide the open source implementation of the IEBSM method \footnote{
    All methods have been implemented with inter alia the \texttt{river} (\hyperref[https://riverml.xyz/]{https://riverml.xyz})  and \texttt{scikit-learn} libraries.}.
     \item We investigate whether statistically significant changes occur in travel mode choice data for a number of travel mode choice data sets and confirm that such changes occured in each of the data sets. Moreover, we confirm that the IEBSM method both detected changes and successfully managed the introduction of selected updated batch models.
\end{itemize}
The remainder of this work is organized as follows. In Sect.~\ref{sec:rel_works}, we provide an overview of related works. This is followed in Sect.~\ref{sec:method} by the proposal of the novel method aiming to automate the use of varied underlying batch and online learners under concept drifting data streams. The results of the evaluation of the method and  reference methods are provided in Sect.~\ref{sec:results}. This is followed by the conclusions and summary of future works in Sect.~\ref{sec:conclusions}.

\section{Related works}
\label{sec:rel_works}
TMC modelling \cite{hillel2021, garcia2022, hagenauer2017} is concerned with predicting the travel mode most likely used by a person for their trip. 
Recently, the  benefits arising from the use of machine learning methods for TMC tasks were discussed in \cite{garcia2022, hagenauer2017}. In \cite{garcia2022}, random forest was shown to yield the best accuracy and computational cost among the tested classifiers. Over time, aspects such as temporal changes in the environment, seasonality, and evolving human preferences are all likely to affect those choices. Hence, in some TMC datasets such as those used in \cite{hagenauer2017}, apart from respondent and trip attributes, such as age, education and distance travelled, the features related to weather conditions were included. Still, batch machine learning methods not considering possible changes in travel mode choice decision boundaries $p(y|x)$ are typically used both in comparative studies~\cite{garcia2022, hagenauer2017} and  surveys of machine learning for TMC modelling \cite{hillel2021}. 

Apart from batch methods, online incremental learning methods have been developed as well, which are also suitable for learning in non-stationary environments \cite{surveyieee}. Notable methods include  
adaptive random forest~\cite{gomes2017adaptive}, which builds upon the random forest method to enable learning from non-stationary data streams. This way, real concept drift \cite{surveyieee}, i.e., a change in $p(y|x)$, can be addressed by changing the ensemble members. In the case of adaptive random forest,  ensemble members can be replaced with new base learners better matching shifted class boundaries $p(y|x)$.
Frequently, the evaluation of online models relies on first making prediction $\hat{y}=h_i(\mathbf{x}_i)$ with the current model $h_i$ to use the instance to get a new, possibly different model $h_{i+1}=learn(h_i,\{\mathbf{x}_i,y_i\})$ through incremental training. This approach is referred to as test-then-train~\cite{moa,gomes2017adaptive}. This illustrates the fact that incremental learning methods respecting stream mining assumptions are constrained by the fact that they can inspect each example at most once~\cite{moa}. This may result in models of a lower performance than the models built within a batch process relying on the access to the entire data set and the ability to iteratively revisit all instances during the training.

As traditional ML models deployed in production settings might experience performance degradation over time, their application to evolving and potentially infinite data streams called for a new approach.
Hence, when Machine Learning solutions are used in IT systems, a growing emphasis on the Machine Learning Operations (MLOps) process is observed. MLOps focuses on addressing data changes through Continuous Monitoring and Continuous Training steps, ensuring models adapt to data and concept shifts to maintain performance
~\cite{testi2022mlops}. %
There are multiple methods for concept drift detection, mainly focusing on
monitoring data  distribution~\cite{moa}. Similarly, the adaptation of batch models can be performed in a number of ways, such as retraining from scratch on all available data~\cite{yang2005line} or just the latest instance window.

The combination of online and batch learning methods has been considered before. In~\cite{grzenda2020hybrid}, neural model training and incremental training of an online learner was proposed in the form of a hybrid model that switches between the multilayer perceptron and stream mining models based on their recent accuracy over a sliding window. 
In~\cite{pishgoo2022dynamic}, the authors  
combined initially trained batch models (e.g. decision trees) and gradually converted them into online models. This approach leverages the simultaneous predictions from both online and batch members.
However, this approach does not involve concept drift detection or model adaptation. Instead, it adds new batch learners built on recent instances over time.

Hence, the question arises of how to build TMC classifiers, while considering concept drift of unknown magnitude. Importantly, not only human preferences towards different modes can change with time (e.g. depend on the time of the year) causing $p(y|x)$ changes, but also $p(x)$ clearly changes. Examples include travel to schools  less likely to happen during school holidays.
\section{Ensemble of Batch and Online Learners}
\label{sec:method}

The method proposed in this work relies on learning an ensemble of base models including both batch  and online models to respond to possible virtual drift, i.e., changes in $p(x)$, and real drift i.e., changes in $p(y|x)$.
To evaluate the IEBSM ensembles as well as reference online and batch methods, the test-then-train approach is applied. Online learners are trained the same way and with the same data stream irrespective of whether they are evaluated as standalone reference learners or participate in the ensemble. Similarly, batch learners are retrained in line with Alg.~\ref{al:comb} described below, both when they are evaluated as reference methods and when they are a part of an ensemble.
\subsection{Training of online and batch learners with TMC data streams}
For online learning algorithms, the learners are provided with new labelled examples and updated incrementally, as defined in the test-then-train approach.
 In the case of batch learning algorithms, 
newly arriving $\{\mathbf{x}_i,y_i\}$ instances are placed in the cache of the most recent instances. Then, drift detectors assess the cache to identify concept drift following the predefined drift handling strategy. If a drift is detected, the batch model undergoes retraining, using data from the cache under the chosen retraining strategy. 
Subsequently, the newly trained model and the previous model are evaluated on the successive $n_{comp}$ instances. If the retrained model demonstrates superior performance compared to the old model, it replaces the previous model in use. This illustrates the challenge of batch learning adapted to an online setting, i.e., its dependence on hyperparameters such as $n_{comp}$.
The batch learning models are trained for the first time using the initial $n_{first\_fit}$ instances. Prior to the collection of all these instances, a majority class model generates prediction output as the label of the class that has been observed most frequently up to that point. In this way, batch and stream models can be evaluated with the same instances $\{\mathbf{x}_i,y_i\},i=1,\dots$ irrespective of the duration of the {\em warm-up} period of $n_{first\_fit}$ instances.%

In our approach, the monitoring strategies used to detect drift rely on the analysis of the most recent instances.
To achieve this, we partition these instances into two equal batches of $s$ instances i.e., the reference batch and the current batch. Every $s$ instances, we then 
compare the distributions of these two batches by applying statistical tests. A test is applied to each feature separately to detect possible changes in the distribution of the values of $j$-th feature $x^j$ and label $y$.
We associate a threshold $\theta$ with those tests, the interpretation of which varies depending on the drift detection method. As the tests to be applied depend on feature types, we discuss them in detail in Sect.~\ref{sec:results}.
Besides testing for changes in the input feature/target distribution, we utilise detection techniques that monitor the performance of a model. A performance drop, defined as the $F_1$ macro score on the current batch falling below $\alpha$ of the $F_1$ macro score on a reference batch, suggests a concept drift. To initiate retraining, at least one drift detection method identifying a change in the data distribution of some feature or a change in model performance must detect a concept drift.

\subsection{Building an ensemble of batch and online learners} \label{sec:comb}
As defined in Alg.~\ref{al:comb}, we propose an ensemble-based method that aims to maximise the performance of travel mode choice predictions by combining predictions of both batch and online learners. The method builds an ensemble of $N$ base learners, some of which can be online learners such as adaptive random forest~\cite{gomes2017adaptive}, while the remaining ones can be batch learners. In line with the test-then-train approach, for every new instance, each base learners generates a prediction first. In the case of batch models, we propose to rely on majority class prediction prior to the collection of a sufficiently large  training data set.

The IEBSM ensemble generates the ultimate prediction by aggregating the outputs from all member models, as shown in lines 16-22 of Alg~\ref{al:comb}. We propose two approaches for combining predictions. In both approaches, we record the value of a performance measure such as $F_1$ for every ensemble member in the prequential approach, i.e., over a sliding window of instances.
    Under the \textbf{Weighted Voting (WV)} approach we combine predictions using weights $w_i$ assigned to each member $m_i$ of ensemble $M$ according to their recent  performance calculated on the sliding window of instances. 
   In the case of the \textbf{Dynamic Switching (DS)} approach, 
    the final prediction of the ensemble is 
    the prediction from the recently top-performing ensemble member. This corresponds to assigning $w_i=1$ to the best model $m_i$, and $w_j=0,j\neq i$ otherwise.

Next the training of base learners is considered. In the case of online learners, they are simply provided with the new instance $(\mathbf{x}_i,y_i)$, which may trigger updates of a model $m$. In the case of batch learners, first a cache of recent instances is updated. This is to store data to be used for potential retraining of a batch model based on a recent data set.
Next, drift handling strategy $S(m)$ is used to define the way drift detection is performed, e.g. whether it is focused on virtual drift only and/or the performance of model $m$, and how sensitive drift detection is, which is defined by the settings of statistical tests. This illustrates the complexity of using batch learners in the case of concept drifting data streams. 

In case a drift is detected, a new model is developed and stored as a shadow model to potentially replace the original one. This happens once its performance is found to be actually superior to the performance of the original model. Hence, in line 33, the new pair of true and predicted labels is used to update the performance of the shadow model $m.shadow\_model$, if any, and compare it to the performance of the original model $m$ and decide whether it should replace the model $m$ or not. In this way, drift detection is combined with the checking of the performance of a newly developed shadow model over a window of new instances not used to train it.
\begin{algorithm}[h]
\caption{Training and evaluation of IEBSM models. \label{al:comb}}
\begin{scriptsize}
\KwIn{$\{x_1, y_2\},\ldots,\{x_i, y_i\},\ldots$ - a labelled data stream, $c$ - a method combining predictions of members, $e$ - a method evaluating members, $S=\{S_1,\ldots,S_K\}, K\leq N$ - a set of drift handling strategies (one per each batch base learner), each defined by a vector of 
hyperparameter values %
controlling the way drift is detected and a batch model retrained}
\KwData{$M=\{m_1,\ldots,m_N\}$ - an ensemble of base learners
}
\ForEach{$\{x_i, y_i\} \in data\_stream$}{
    $\hat{Y}_i \gets [\space \space]$, $scores_i \gets [\space \space]$ \\ 
    \ForEach{$m \in M$}{
        \If {$m.type == batch$ \space \textbf{and} \space $i <= n_{first\_fit}$}
        {
            $\hat{Y}_{i}[m] \gets get\_majority\_class(m.cache)$ \\
            \If {$i == n_{first\_fit}$}
            {
                $m = m.first\_fit()$
            }
        } 
        \Else
        {
            $\hat{Y}_{i}[m] \gets m.predict(x_i)$ \\
        }
        
        $e.update\_model\_score(y_i,\hat{Y}_{i}[m]) $\\
    }
    \If{$c == DS$}
    {   
        $weigths \gets zeros(len(members))$\\
        $weights[argmax(scores_i)] \gets 1$
    }
    \If{$c == WV$}
    {
        $weights \gets (scores_i / sum(scores_i))$
    }
    
    $\hat{y}_i \gets argmax_{c \in classes}\{{\Sigma_{j\gets0, \hat{Y}_{i}[j] == c}^{m_{len} - 1}{weights[j]}\}}$\\
    \ForEach{$m \in M$}{
        \If{$m.type == online$}
        {
            $m=m.update\_model(x_i, y_i)$\\
        }
        \If{$m.type == batch$}
        {
            
            $m.update\_cache(x_i, y_i)$\\
            \If{$m.has\_concept\_drift\_occurred(S(m))$}
            {
                $m.shadow\_model \gets train\_on\_cached\_instances(m, S(m))$\\
                
            }
             $m \gets evaluate\_shadow\_model(m,m.shadow\_model,x_i, y_i,\hat{Y}_{i}[m])$\\
        }
    }
    $update\_performance\_metrics(y_i, \hat{y}_{i})$\\
}
\end{scriptsize}

\end{algorithm}

\begin{table}[h]
\centering
\caption{The summary of data streams}
\begin{tabular}{|c|c|c|c|p{6cm}|}
\hline
Data stream & Instances & Features & Classes & Description \\
\hline
Ohio (OHI) & 122,331 & 156 & 12 & 2001-2003 Ohio survey ~\cite{team20232001}\\
\hline
London (LON) & 81,086 & 41 & 4 &  London Travel Demand 
Survey~\cite{hillel2018recreating}\\
\hline
Optima (OPT) & 2,265 & 497 & 4 & Suiss survey data~\cite{bierlaire2018mode}\\
\hline
NTS & 230,608 & 17 & 4 & Dutch National Travel Survey with environment and weather features ~\cite{garcia2022, hagenauer2017} \\
\hline
N-MW & 144,905 & 2,571 & 21 &  The National  Household Travel\\
N-NE & 145,564 & 2,437 & 21 & Survey (NHTS) conducted in \\
N-SE & 209,485 & 2,586 & 21 & 2016 and 2017~\cite{USDOT2017}, divided\\
N-SW & 190,279 & 2,505 & 21 &  into five regions of the US \\
N-W & 233,323 & 2,553 & 21 &  \\
\hline
\end{tabular}
\label{tab:data}
\end{table}

\section{Results}
\label{sec:results}

\subsection{Data streams and libraries}

The experiments performed with online, batch, and combined methods were assessed on real travel mode data streams, with the overview provided in Table~\ref{tab:data}. The datasets vary in the number of features, classes, i.e., travel modes, and instances.
Each instance corresponds to an actual trip reported by a survey participant, with the employed travel mode designated as the target variable. For a more comprehensive description of data stream preparation, please refer to the supplementary material, where additional description is provided.

To implement the proposed methods and the evaluation framework, we  used the library \textsf{River} ~\cite{montiel2021river} (online learning methods),
and the \textsf{LightGBM} package (the LGBM classifier). For the batch learning methods, except for LGBM,
the \textsf{scikit-learn} library
was used, while %
the concept drift detection implementation relied on the \textsf{EvidentlyAI}~\cite{Evidentl19:online}.

\subsection{Experiments}
We conducted a series of experiments for each data stream, which included online learning experiments, batch learning experiments, and experiments using the IEBSM method.
The precise configurations of online and batch learning methods are detailed in the supplementary material accompanying this work.

For online learning, we employed  the Hoeffding Adaptive Tree (HAT), Adaptive Random Forest (ARF), Streaming Random Patches (SRP), Online Gaussian Naive Bayes ((O)NB), and Online Logistic Regression ((O)LR) algorithms.
The batch learning experiments made use of the Logistic Regression (LR) (with prior standardization of the training batch), Gaussian Naive Bayes (NB), Decision Tree Classifier (DT), LGBM, and Random Forest (RF) algorithms.
Furthermore, we applied each batch learning algorithm to the data streams using three distinct drift handling strategies:
\begin{itemize}
    \item S1: Basic drift detection of changes in input features, target and model performance drift with $\theta=0.03$, $s=10,000$, $\alpha=0.2$
    \item S2: Performance drift detection only with $s=10,000$, $\alpha=0.2$
    \item S3: Frequent drift detection of changes in input features, target and model performance with $\theta=0.02$, $s=5,000$, $\alpha=0.2$, i.e., relying on smaller windows of $5,000$ instances than in S1.
\end{itemize}
Batch experiments were equivalent to running Alg.~\ref{al:comb} with one batch base learner and its drift handling strategy.
We have chosen the values for the hyperparameters $\theta$ and $s$ through preliminary tests aimed at determining the values resulting in possibly high performance of the models. In the batch-learning experiments, we employed a retraining strategy, which involved training a new model using all historical instances that arrived after the last model replacement.
After each retraining, we assessed the performance of a shadow model relative to the old one, over the following $n_{comp}=500$ instances. Depending on their performance, we would replace the old model with the new one. 
The first training took place after the initial
$n_{first\_fit}=2500$ examples. %

Moreover, we conducted baseline experiments in which we trained each batch algorithm on the initial 2,500 instances (strategy B1) 
and subsequently used that model for predictions on the remaining data stream.
We also repeated the baseline experiments using an initial training set of 25,000 instances (strategy B2) for a more comprehensive analysis.

Finally, in the experiments including batch models and involving drift detection and model adaptation we dynamically selected a specific statistical test based on 
the input feature/target column.
For numerical columns with the number of unique values $n\_unique > 5$ we used \textsf{Wasserstein Distance} when $s > 1,000$; and two-sample \textsf{Kolmogorov-Smirnov} test otherwise.
For categorical columns or numerical (with $2 < n\_unique <= 5$), \textsf{Jensen--Shannon} divergence when $s > 1,000$; or \textsf{chi-squared} test were used otherwise.
Finally, for binary categorical features ($n\_unique = 2$):  \textsf{Jensen--Shannon} divergence was used when $s > 1,000$; and proportion difference test for independent samples based on $Z$-score otherwise.

\subsubsection{Combining online and batch learning with the IEBSM method}
The  experiments using IEBSM included an ensemble of four instances of the same batch learning classifier, each utilizing a distinct drift handling (DH) strategy, along with three online learning classifiers: (O)NB, HAT, and ARF. We tested the LGBM and RF as the batch learning algorithms. We employed a single batch learning algorithm for all batch members within each IEBSM experiment to reduce variation arising from diverse algorithms. This allowed us to single out the impact of distinct drift handling strategies, namely:
\begin{itemize}
    \item S4: investigating changes in input features, target, and model performance  with $\theta=0.02$, $s=2,500$, $\alpha=0.2$
    \item S5: investigating changes in model performance only  with $s=2,500$, $\alpha=0.2$
    \item S6: investigating changes in input features, target, and model 
    performance with $\theta=0.03$, $s=10,000$, $\alpha=0.2$
    \item S7: investigating changes  in input features, target, and model performance with $\theta=0.02$, $s=10,000$, $\alpha=0.2$
\end{itemize}
In the S5 and S7 settings, the retraining batch corresponds to the window of last $s$ instances.
In contrast, for S4 and S6, all instances since the last model replacement are considered for retraining. The other hyperparameter values (e.g., $n_{first\_fit}, n_{comp})$ were the same as in the single online/batch learning experiments.
The seven ensemble members described above were combined using two methods $c$ outlined in Section \ref{sec:comb}.

After conducting experiments, it became evident that online learning methods exhibited significantly inferior performance compared to batch learning methods. To demonstrate the effect of model combination with the IEBSM method and eliminate the impact of under-performing online models, we conducted additional IEBSM experiments: DS-BATCH and WV-BATCH experiments utilizing only LGBM S4, S5, S6, and S7 models. Moreover, we performed DS-ONLINE and WV-ONLINE experiments combining HAT, ARF, and (O)NB models solely.
For the nine data streams, we calculated each method's average ranking position based on the value of $F_1$ macro score. Table \ref{tab:res} shows the obtained results for the selected experiments (all experiment results are provided in the supplementary material).

\begin{table}
\centering
\small
\caption{Ranks of selected methods across all streams and $F_1$ macro score for each data stream. %
Data streams were arranged in order based on the increasing number of features. A ranking score combined with the corresponding position in the overall ranking (in brackets). $\dagger$ - For the OPT data stream, hyperparameters values set to $s = 100$ and $s = 250$ (instead of 2,500 and 10,000), $n_{first\_fit}=150$, and $n_{comp}=50$. For N-* data streams,   the input feature drift detection disabled in DH strategies S1, S3, S4, S6, S7 due to the performance issues caused by a large number of features.}
\begin{tabular}{|cc|ccccccccc|}
\hline
Method &Rank& \multicolumn{9}{c|}{\textit{Data stream 
}} \\ 
 & (pos.) & NTS & LON & OHI& OPT$\dagger$ & N-NE & N-SW & N-W & N-MW & N-SE \\
\hline
        DS-RF &  4.33(1) & 0.532 &   0.544 & 0.206 &   0.393 &    \textbf{0.464} &    \textbf{0.464} &   \textbf{0.453} &    \textbf{0.476} &    0.453 \\
        WV-RF &  5.78(2) & 0.534 &   0.546 & 0.197 &   0.362 &    0.460 &    0.460 &   0.435 &    0.458 &    \textbf{0.456} \\
      DS-LGBM &  7.78(3) & 0.540 &   \textbf{0.549} & \textbf{0.224} &   0.436 &    0.377 &    0.358 &   0.320 &    0.368 &    0.316 \\
        RF-S3 &  8.11(4) & 0.520 &   0.530 & 0.205 &   0.382 &    0.414 &    0.421 &   0.417 &    0.403 &    0.428 \\
DS-BATCH &  8.44(5) & 0.541 &   0.538 & \textbf{0.224} &   0.446 &    0.379 &    0.357 &   0.320 &    0.365 &    0.314 \\
      WV-LGBM &  9.00(6) & \textbf{0.542} &   0.546 & 0.215 &   0.438 &    0.359 &    0.357 &   0.306 &    0.352 &    0.316 \\
WV-BATCH & 10.83(12) & 0.539 &   0.531 & 0.216 &   \textbf{0.459} &    0.358 &    0.356 &   0.307 &    0.349 &    0.314 \\
      LGBM-S3 & 12.39(13) & 0.530 &   0.533 & 0.213 &   0.453 &    0.358 &    0.339 &   0.302 &    0.349 &    0.314 \\
      LGBM-B1 & 18.61(19) & 0.355 &   0.512 & 0.213 &   0.307 &    0.358 &    0.339 &   0.302 &    0.349 &    0.314 \\
        RF-B2 & 23.11(23) & 0.486 &   0.426 & 0.196 &   0.274 &    0.279 &    0.229 &   0.197 &    0.232 &    0.220 \\
      LGBM-B2 & 25.00(24) & 0.506 &   0.433 & 0.162 &   0.299 &    0.167 &    0.179 &   0.183 &    0.118 &    0.142 \\
    WV-ONLINE & 25.44(26) & 0.481 &   0.504 & 0.166 &   0.338 &    0.077 &    0.071 &   0.066 &    0.081 &    0.069 \\
          SRP & 26.22(27) & 0.375 &   0.430 & 0.177 &   0.234 &    0.231 &    0.161 &   0.210 &    0.210 &    0.195 \\
        RF-B1 & 26.33(28) & 0.356 &   0.516 & 0.164 &   0.199 &    0.253 &    0.211 &   0.146 &    0.189 &    0.167 \\
    DS-ONLINE & 26.78(29) & 0.375 &   0.432 & 0.162 &   0.312 &    0.084 &    0.080 &   0.074 &    0.090 &    0.078 \\
\hline
\end{tabular}
\label{tab:res}
\vspace{-0.5cm}
\end{table}

\subsection{Discussion}
It follows from Table~\ref{tab:res} that  batch-learning experiments RF-S3 and LGBM-S3 employing DH strategies to possibly adapt batch models, outperformed the baseline experiments RF-B* and LGBM-B* in which RF and LGBM models trained once on initial batch of data were used next to predict travel modes for all the remaining instances (B1 and B2 strategies, sample results in Table~\ref{tab:res} provided inter alia for RF as RF-B1 results). 
This finding demonstrates that implementing the aforementioned DH strategies significantly benefits batch-learning models when faced with travel mode choice data. 
Among different ways the RF and LGBM models can be updated, strategy S3 stood out as the most effective.
\par
Surprisingly, the online learning methods yielded the poorest performance results, as illustrated by the SRP results. One potential explanation could be the abundance of features in our data streams, among which many might be irrelevant. The SRP classifier emerged as the most effective online learning method, albeit with the trade-off of longer execution times.
\par
The use of IEBSM, evaluated in the DS-* and WV-* experiments,  enhanced performance compared to individual batch and online model setups, and provided the best overall results. This combination successfully mitigated the challenges caused by the need to choose online and batch learning methods. Moreover, utilising multiple batch members with varied DH strategies and varied hyperparameters in turn reduced the need to pre-select the optimal strategy. While the Random Forest algorithm used to develop batch ensemble members yielded superior results, as shown by the DS-RF outcome,  the DS-LGBM demonstrated faster operation despite the inferior performance.
\par
Interestingly, the IEBSM ensembles comprised solely of batch-learning LGBM classifiers ([DS/WV]-BATCH) resulted in a worse global rank than the corresponding experiments that combined both LGBM and online classifiers ([DS/WV]-LGBM).
These findings underscore that, even in cases where online learning demonstrates suboptimal performance, IEBSM ensembles can benefit from the diversity their members offer. Similarly, when using combining ensembles solely with online learners ([DS/WV]-ONLINE), a notable enhancement in performance, compared to the performance of the experiments utilizing single instances of HAT, ARF, and O(NB), was observed.

\begin{figure}[t]
\begin{center}
\caption{The number of drift detections and model replacements. DS-RF experiments.}
\includegraphics[width=0.7\textwidth]{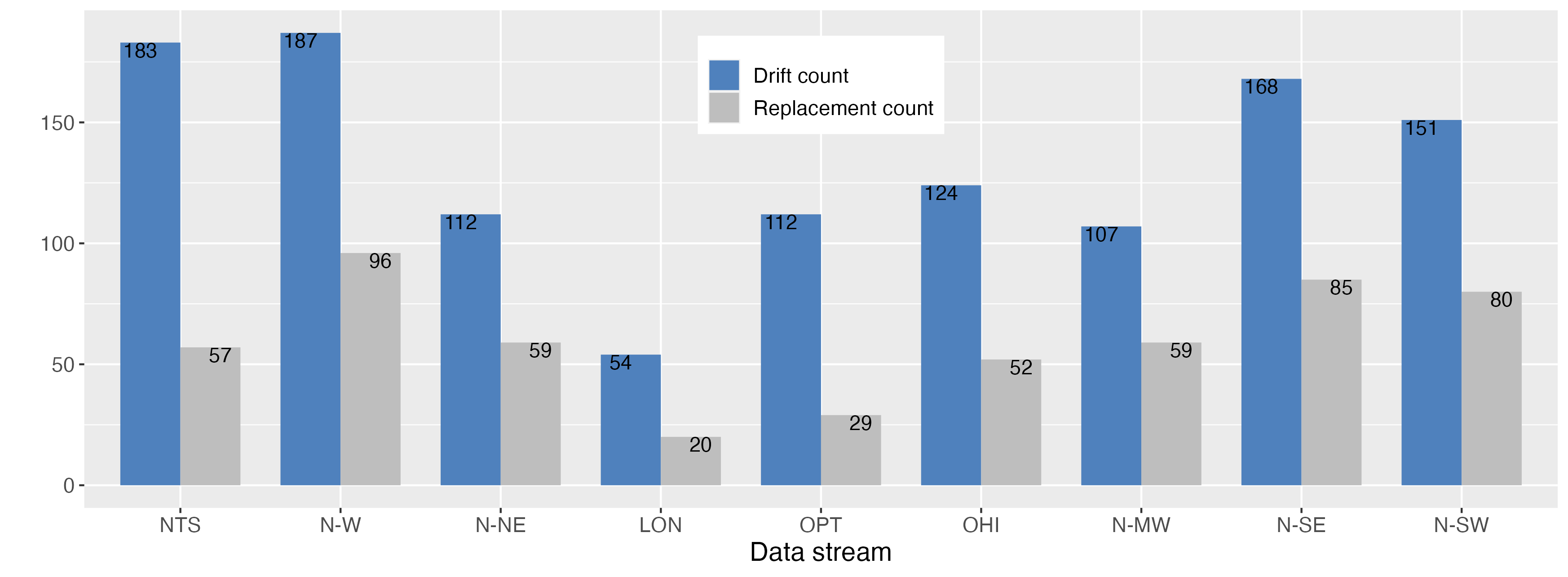}
\label{fig:det}
\end{center}
\vspace{-1.0cm}
\end{figure}

Finally, the role of drift detections and shadow models in the IEBSM approach can be analyzed. 
Fig.~\ref{fig:det} presents the total number of drift detections and actual model replacements for all batch members for the  highest-ranked method, i.e., the IEBSM-based DS-RF~\footnote{Due to the extensive number of experiments conducted, detailed results  for all experiments are provided in the supplementary materials.} ensemble, which notably included both online and random forest models.
It follows from the figure, that detections of statistically significant changes in data have occurred in all data streams. These were followed  by the replacement of batch models. Hence, the shadow models built with more recent data were found to be superior to the original models they replaced. 
Newly developed models were only in some cases found to yield better performance, as the drift count significantly exceeds the actual model replacement count. This confirms that both detection and the evaluation of shadow models are vital components of the highest-ranked approach to building TMC models i.e., the DS-RF approach. 
\section{Conclusions}
\label{sec:conclusions}
In the prevailing majority of cases, modelling of travel mode choices is performed with batch learning methods. However, factors such as seasonality suggest that when predicting TMC decisions
incorporating concept drift detection and adaptation could be justified. On the other hand, change detection could occur too frequently and reduce the potential of newly developed models in turn. A possible solution to the problem can rely on the use of both online and batch learners.
Our experiments performed with multiple travel mode choice data sets  confirm the need for continuous monitoring and retraining of TMC models. Combining batch and online learning clearly yields improved  performance of the models. Furthermore, the IEBSM  method  %
eases the challenge of choosing a learning method and drift detection settings
by employing multiple base members including both online and batch learners with different drift handling strategies. This resulted in the best rank of the IEBSM approach.

Future works entail exploring various combining approaches, such as different ways of assigning member weights. Furthermore, travel mode choice data sets can be used to foster the development of future stream mining methods.

\subsubsection*{Acknowledgements}{
Paweł Golik:
The reasearch has been conducted as part of the third edition of the CyberSummer@WUT-3 competition
at
the Warsaw University of Technology as part of the Research Centre POB for Cybersecurity
and Data Science (CB POB Cyber\&DS).
Maciej Grzenda: The research leading to these results has received funding from the EEA/Norway Grants 2014-2021 through the National Centre for Research and Development. CoMobility benefits from a 2.05 million \texteuro grant from Iceland, Liechtenstein and Norway through the EEA Grants. The aim of the project is to provide a package of tools and methods for the co-creation of sustainable mobility in urban spaces.


\begin{thebibliography}{10}
\providecommand{\url}[1]{\texttt{#1}}
\providecommand{\urlprefix}{URL }
\providecommand{\doi}[1]{https://doi.org/#1}

\bibitem{Evidentl19:online}
Evidently ai - open-source machine learning monitoring. \url{https://www.evidentlyai.com/}, (Accessed on 05/10/2023)

\bibitem{bierlaire2018mode}
Bierlaire, M.: Mode choice in {Switzerland (Optima)}. URL: https://transp-or. epfl. ch/documents/technicalReports/CS\_OptimaDescription. pdf  (2018)

\bibitem{moa}
Bifet, A., Gavaldà, R., Pfahringer, B., Holmes, G.: Machine Learning for Data Streams with Practical Examples in MOA. MIT Press (03 2018)

\bibitem{surveyieee}
Ditzler, G., Roveri, M., Alippi, C., Polikar, R.: Learning in nonstationary environments: A survey. IEEE Computational Intelligence Magazine  \textbf{10}(4),  12--25 (2015)

\bibitem{gama2010knowledge}
Gama, J.: Knowledge discovery from data streams. CRC Press (2010)

\bibitem{gandomani2020role}
Gandomani, T.J., Tavakoli, Z., Zulzalil, H., Farsani, H.K.: The role of project manager in agile software teams: A systematic literature review. IEEE Access  \textbf{8},  117109--117121 (2020)

\bibitem{garcia2022}
García-García, J.C., García-Ródenas, R., López-Gómez, J.A., Ángel Martín-Baos, J.: A comparative study of machine learning, deep neural networks and random utility maximization models for travel mode choice modelling. Transportation Research Procedia  \textbf{62},  374--382 (2022)

\bibitem{gomes2017adaptive}
Gomes, H.M., Bifet, A., Read, J., Barddal, J.P., Enembreck, F., Pfharinger, B., Holmes, G., Abdessalem, T.: Adaptive random forests for evolving data stream classification. Machine Learning  \textbf{106}(9),  1469--1495 (2017)

\bibitem{grzenda2020hybrid}
Grzenda, M., Kwasiborska, K., Zaremba, T.: Hybrid short term prediction to address limited timeliness of public transport data streams. Neurocomputing  \textbf{391},  305--317 (2020)

\bibitem{hagenauer2017}
Hagenauer, J., Helbich, M.: A comparative study of machine learning classifiers for modeling travel mode choice. Expert Systems with Applications  \textbf{78},  273--282 (7 2017)

\bibitem{hillel2021}
Hillel, T., Bierlaire, M., Elshafie, M.Z., Jin, Y.: A systematic review of machine learning classification methodologies for modelling passenger mode choice. Journal of Choice Modelling  \textbf{38},  100221 (2021)

\bibitem{hillel2018recreating}
Hillel, T., Elshafie, M.Z., Jin, Y.: Recreating passenger mode choice-sets for transport simulation: A case study of {London, UK}. Proceedings of the Institution of Civil Engineers-Smart Infrastructure and Construction  \textbf{171}(1),  29--42 (2018)

\bibitem{montiel2021river}
Montiel, J., Halford, M., Mastelini, S.M., Bolmier, G., Sourty, R., Vaysse, R., Zouitine, A., Gomes, H.M., Read, J., Abdessalem, T., et~al.: River: machine learning for streaming data in python  (2021)

\bibitem{pishgoo2022dynamic}
Pishgoo, B., Azirani, A.A., Raahemi, B.: A dynamic feature selection and intelligent model serving for hybrid batch-stream processing. Knowledge-Based Systems  \textbf{256},  109749 (2022)

\bibitem{team20232001}
Team, T.: 2001--2003 ohio statewide household travel survey. Tech. rep., Livewire Data Platform; NREL; Pacific Northwest National Lab.(PNNL) (2023)

\bibitem{testi2022mlops}
Testi, M., Ballabio, M., Frontoni, E., Iannello, G., Moccia, S., Soda, P., Vessio, G.: Mlops: A taxonomy and a methodology. IEEE Access  (2022)

\bibitem{USDOT2017}
{U.S. Department of Transportation, Federal Highway Administration}: 2017 national household travel survey. URL: \url{http://nhts.ornl.gov} (2017)

\bibitem{yang2005line}
Yang, J., Rivard, H., Zmeureanu, R.: On-line building energy prediction using adaptive artificial neural networks. Energy and buildings  \textbf{37}(12),  1250--1259 (2005)

\end{thebibliography}
\end{document}


\maketitle

\section{Global ranking}
Table \ref{tab:rank} illustrates the comprehensive ranking of methods. Across the 9 data streams, all 38 methods were organized based on their $F_1$ macro scores in descending order. Subsequently, we computed the \textit{Ranking score} for each method by computing their average ranking position across these nine rankings. The ultimate \textit{Ranking position} represents the sequential number of these averages, sorted from the smallest to the nearest value.

\pagebreak

\begin{longtable}{llr|llr}
\caption{Table with the global ranking of used methods} \\
\label{tab:rank} \\
\midrule
{Ranking position} & Method abbr. & Ranking score & {Ranking position} & Method abbr. & Ranking score \\
\midrule
\endhead
1  &          DS-RF &   4.33 & 20 &          DT B2 &  19.56 \\
2  &          WV-RF &   5.78 & 21 &          DT B1 &  21.67 \\
3  &        DS-LGBM &   7.78 & 22 &          LR B2 &  22.11 \\
4  &          RF S3 &   8.11 & 23 &          RF B2 &  23.11 \\
5  &  DS-BATCH &   8.44 & 24 &        LGBM B2 &  25.00 \\
6  &        WV-LGBM &   9.00 & 25 &          LR B1 &  25.11 \\
7  &          RF S1 &   9.00 & 26 &      WV-ONLINE &  25.44 \\
8  &          LR S3 &   9.78 & 27 &            SRP &  26.22 \\
9  &          DT S1 &   9.83 & 28 &          RF B1 &  26.33 \\
10  &          DT S3 &  10.28 & 29 &      DS-ONLINE &  26.78 \\
11 &          LR S1 &  10.33 & 30 &          NB S1 &  30.44 \\
12 &  WV-BATCH &  10.83 & 31 &            HAT &  30.78 \\
13 &        LGBM S3 &  12.39 & 32 &          NB S3 &  31.22 \\
14 &        LGBM S1 &  12.56 & 33 &          NB B2 &  31.78 \\
15 &          RF S2 &  13.06 & 34 &            ARF &  32.00 \\
16 &        LGBM S2 &  13.28 & 35 &          NB S2 &  32.50 \\
17 &          LR S2 &  14.00 & 36 &          NB B1 &  34.39 \\
18 &          DT S2 &  16.39 & 37 &            ONB &  35.56 \\
19 &        LGBM B1 &  18.61 & 38 &            OLR &  37.22 \\
\bottomrule
\end{longtable}

\section{Data stream preparation}
If any of the nine data streams included variables related to the date and time of the journey, the instances were arranged chronologically.
In each original data stream, we removed variables that might lead to knowledge leakage and conducted one-hot encoding for categorical variables. The datasets had minimal missing values, and for categorical variables, we converted these to a category indicating \textit{'Don't know / Refuse to answer'}. Numerical missing values were replaced with the mode value computed across the entire dataset. Instances with missing target values were excluded.

\section{Online and batch learning models configuration}
All batch learning models were initialized with their default hyperparameter values, except for setting the \textit{random\_seed} to 42 where applicable. Within Listings \ref{listing:1} to \ref{listing:5}, you'll find code snippets that define online learning models using the \textsf{River} library.

\newpage
\begin{lstlisting}[language=Python, caption=Online Logistic Regression (OLR) model definition,label={listing:1}]
from river.linear_model import LogisticRegression as LROnline
from river import compose
from river.preprocessing import StandardScaler
from river import optim
 
lr_online = compose.Pipeline(
    StandardScaler(
        with_std=True
    ),
    LROnline(
        optimizer=optim.SGD(
            lr=0.005
        ),
        loss=optim.losses.Log(
            weight_pos=1.,
            weight_neg=1.
        ),
        l2=1.0,
        l1=0.,
        intercept_init=0.,
        intercept_lr=0.01,
        clip_gradient=1e+12,
        initializer=optim.initializers.Zeros()
    )
)
\end{lstlisting}

\begin{lstlisting}[language=Python, caption=Adaptive Random Forest (ARF) model definition,label={listing:2}]
from river import forest

arf = forest.ARFClassifier(seed=42, leaf_prediction="mc")
\end{lstlisting}

\begin{lstlisting}[language=Python, caption=Hoeffding Adaptive Tree (HAT) model definition,label={listing:3}]
from river.tree import HoeffdingAdaptiveTreeClassifier
 
hat = HoeffdingAdaptiveTreeClassifier(
    grace_period=100,
    delta=0.01,
    leaf_prediction='nb',
    nb_threshold=10,
    seed=42
)
\end{lstlisting}

\begin{lstlisting}[language=Python, caption=Streaming Random Patches (SRP) model definition,label={listing:4}]
from river.tree import HoeffdingTreeClassifier
from river import ensemble

base_model = HoeffdingTreeClassifier(grace_period=100, delta=0.01)
srp_model = ensemble.SRPClassifier(model=base_model, n_models=3, seed=42)
\end{lstlisting}
\newpage
\begin{lstlisting}[language=Python, caption=Online Gaussian Naive Bayes (ONB) model definition,label={listing:5}]
from river.naive_bayes import GaussianNB as GNBOnline
 
nb_online = GNBOnline()
\end{lstlisting}

\newpage

\section{Detailed results}
\begin{figure}[]
\begin{center}
\caption{Visualization of $F_1$ macro score for selected methods.}
\includegraphics[height=0.99\textheight]{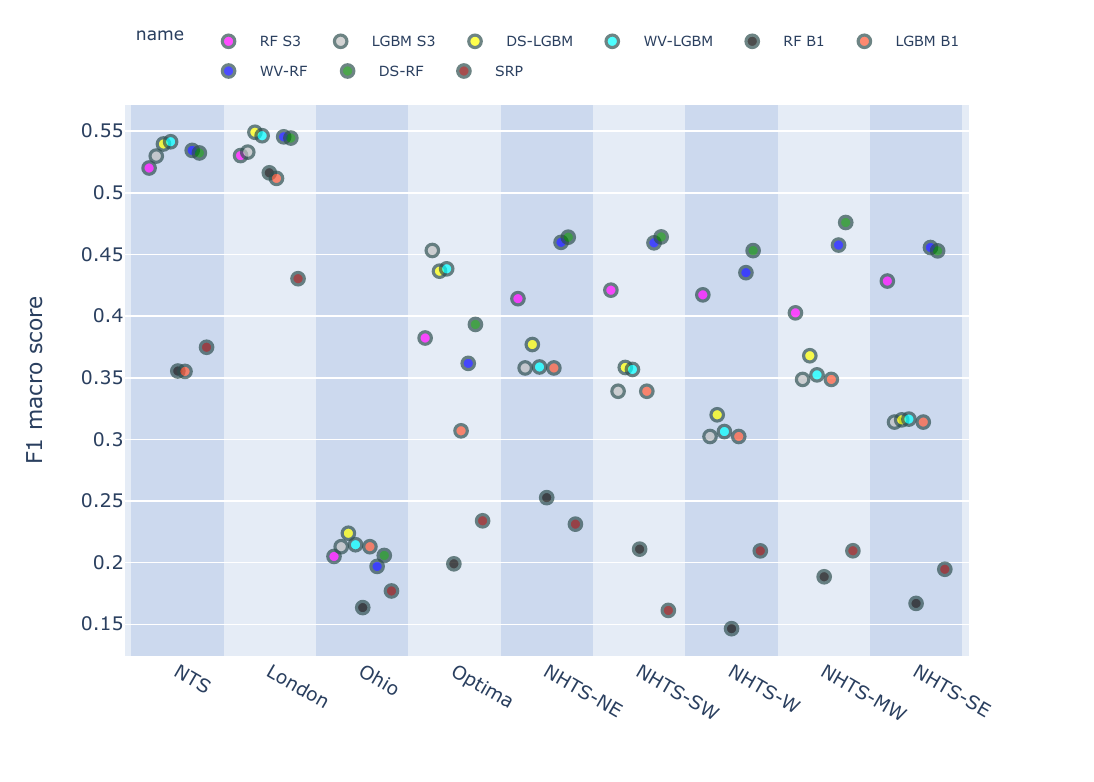}
\label{fig:res}
\end{center}
\end{figure}

Within Table \ref{tab:1}, you'll find a comprehensive breakdown of experiment outcomes across various data streams. The rows are arranged based on both the data stream and $F_1$ macro score. For ensembles, the presented drift/replacement values represent the aggregated sum across all ensemble members. Additionally, with respect to both online and baseline methods, the count of drifts and replacements is zero since these methods do not utilize our monitoring and retraining strategies. Figure \ref{fig:res} presents the $F_1$ macro score values for selected methods on all data streams. The data streams were arranged in order based on the increasing number of features.

\pagebreak

\pagebreak

\begin{longtable}
{llllrrrrrrrrr}
\caption{Table with detailed results for all experiments} \\
\label{tab:1} \\
\hline
{} &  Data stream &  Global rank pos. & Method abbr. &  $F_1$ macro &  Accuracy &     Kappa &  Drift count &  Replacement count & Time [s] \\
\hline
\endfirsthead

\hline
{} &  Data stream &  Global rank pos. & Method abbr. & $F_1$ macro &  Accuracy &     Kappa &  Drift count &  Replacement count & Time [s] \\
\hline
\endhead

\hline
\endfoot

\endlastfoot
0   &   London &      3 &        DS-LGBM &  0.5492 &  0.7347 &  0.5830 &   45.0 &       17.0 &    1083.41 \\
1   &   London &      6 &        WV-LGBM &  0.5464 &  0.7339 &  0.5838 &   45.0 &       17.0 &    1045.94 \\
2   &   London &      2 &          WV-RF &  0.5455 &  0.7317 &  0.5804 &   54.0 &       20.0 &    5789.70 \\
3   &   London &      1 &          DS-RF &  0.5445 &  0.7331 &  0.5795 &   54.0 &       20.0 &    5778.41 \\
4   &   London &      5 &  DS-BATCH &  0.5378 &  0.7198 &  0.5633 &   22.0 &       11.0 &     583.30 \\
5   &   London &     14 &        LGBM S1 &  0.5332 &  0.7157 &  0.5580 &   15.0 &        8.0 &     143.40 \\
6   &   London &     13 &        LGBM S3 &  0.5331 &  0.7180 &  0.5606 &    6.0 &        2.0 &     125.21 \\
7   &   London &     16 &        LGBM S2 &  0.5324 &  0.7144 &  0.5552 &    1.0 &        1.0 &     100.41 \\
8   &   London &     12 &  WV-BATCH &  0.5306 &  0.7151 &  0.5561 &   22.0 &       11.0 &     499.85 \\
9   &   London &      4 &          RF S3 &  0.5303 &  0.7156 &  0.5578 &    6.0 &        1.0 &    1608.12 \\
10  &   London &     11 &          LR S1 &  0.5299 &  0.6943 &  0.5312 &   14.0 &        9.0 &     149.13 \\
11  &   London &      7 &          RF S1 &  0.5281 &  0.7137 &  0.5532 &   14.0 &        9.0 &    1579.17 \\
12  &   London &      8 &          LR S3 &  0.5274 &  0.7065 &  0.5446 &    8.0 &        2.0 &     158.33 \\
13  &   London &     17 &          LR S2 &  0.5163 &  0.6811 &  0.5124 &    1.0 &        1.0 &      84.32 \\
14  &   London &     15 &          RF S2 &  0.5163 &  0.6951 &  0.5295 &    0.0 &        0.0 &    1458.65 \\
15  &   London &     28 &          RF B1 &  0.5163 &  0.6951 &  0.5295 &    0.0 &        0.0 &    1231.51 \\
16  &   London &     25 &          LR B1 &  0.5160 &  0.6808 &  0.5119 &    0.0 &        0.0 &      51.99 \\
17  &   London &     19 &        LGBM B1 &  0.5118 &  0.6921 &  0.5217 &    0.0 &        0.0 &    1278.48 \\
18  &   London &     26 &      WV-ONLINE &  0.5042 &  0.6584 &  0.4892 &    0.0 &        0.0 &     536.18 \\
19  &   London &     10 &          DT S3 &  0.4534 &  0.5853 &  0.3696 &    6.0 &        5.0 &      98.58 \\
20  &   London &      9 &          DT S1 &  0.4534 &  0.5858 &  0.3708 &   14.0 &        5.0 &      89.98 \\
21  &   London &     21 &          DT B1 &  0.4429 &  0.5645 &  0.3440 &    0.0 &        0.0 &      38.39 \\
22  &   London &     18 &          DT S2 &  0.4411 &  0.5650 &  0.3425 &    0.0 &        0.0 &      53.19 \\
23  &   London &     24 &        LGBM B2 &  0.4333 &  0.5718 &  0.3819 &    0.0 &        0.0 &    1102.58 \\
24  &   London &     34 &            ARF &  0.4319 &  0.7292 &  0.5698 &    0.0 &        0.0 &     383.62 \\
25  &   London &     29 &      DS-ONLINE &  0.4317 &  0.7278 &  0.5685 &    0.0 &        0.0 &     587.02 \\
26  &   London &     27 &            SRP &  0.4304 &  0.7092 &  0.5424 &    0.0 &        0.0 &     899.10 \\
27  &   London &     30 &          NB S1 &  0.4292 &  0.5202 &  0.3408 &   14.0 &        7.0 &     109.38 \\
28  &   London &     23 &          RF B2 &  0.4264 &  0.5672 &  0.3748 &    0.0 &        0.0 &    1072.53 \\
29  &   London &     22 &          LR B2 &  0.4248 &  0.5646 &  0.3717 &    0.0 &        0.0 &      49.88 \\
30  &   London &     32 &          NB S3 &  0.4236 &  0.5125 &  0.3323 &    6.0 &        4.0 &     132.09 \\
31  &   London &     35 &          NB S2 &  0.4194 &  0.5110 &  0.3305 &    0.0 &        0.0 &      78.13 \\
32  &   London &     36 &          NB B1 &  0.4194 &  0.5110 &  0.3305 &    0.0 &        0.0 &      52.06 \\
33  &   London &     20 &          DT B2 &  0.3840 &  0.4856 &  0.2695 &    0.0 &        0.0 &      37.40 \\
34  &   London &     37 &            ONB &  0.3545 &  0.5518 &  0.3650 &    0.0 &        0.0 &     125.82 \\
35  &   London &     33 &          NB B2 &  0.3523 &  0.4226 &  0.2356 &    0.0 &        0.0 &      49.99 \\
36  &   London &     31 &            HAT &  0.3385 &  0.5092 &  0.3168 &    0.0 &        0.0 &     319.72 \\
37  &   London &     38 &            OLR &  0.0208 &  0.0319 &  0.0020 &    0.0 &        0.0 &      54.64 \\
\midrule
38  &  NHTS-MW &      1 &          DS-RF &  0.4760 &  0.7497 &  0.6402 &  107.0 &       59.0 &   84720.46 \\
39  &  NHTS-MW &      2 &          WV-RF &  0.4577 &  0.7435 &  0.6284 &  107.0 &       59.0 &   67835.22 \\
40  &  NHTS-MW &      7 &          RF S1 &  0.4424 &  0.7299 &  0.6133 &   34.0 &       16.0 &    3796.36 \\
41  &  NHTS-MW &      8 &          LR S3 &  0.4237 &  0.5872 &  0.4435 &   11.0 &       10.0 &    1089.48 \\
42  &  NHTS-MW &     11 &          LR S1 &  0.4161 &  0.6145 &  0.4713 &   31.0 &       19.0 &    1420.77 \\
43  &  NHTS-MW &      9 &          DT S1 &  0.4105 &  0.7464 &  0.6559 &   32.0 &       20.0 &     601.91 \\
44  &  NHTS-MW &     10 &          DT S3 &  0.4036 &  0.7330 &  0.6374 &   11.0 &        9.0 &     281.73 \\
45  &  NHTS-MW &     15 &          RF S2 &  0.4027 &  0.7137 &  0.5873 &   10.0 &        9.0 &    3312.46 \\
46  &  NHTS-MW &      4 &          RF S3 &  0.4027 &  0.7137 &  0.5873 &   12.0 &        9.0 &    2548.20 \\
47  &  NHTS-MW &     17 &          LR S2 &  0.3858 &  0.5747 &  0.4308 &    7.0 &        7.0 &     818.72 \\
48  &  NHTS-MW &      3 &        DS-LGBM &  0.3679 &  0.7605 &  0.6610 &   48.0 &        1.0 &   65638.91 \\
49  &  NHTS-MW &      5 &  DS-BATCH &  0.3651 &  0.7533 &  0.6536 &   58.0 &        1.0 &    8220.94 \\
50  &  NHTS-MW &      6 &        WV-LGBM &  0.3524 &  0.7537 &  0.6507 &   48.0 &        1.0 &   71879.20 \\
51  &  NHTS-MW &     16 &        LGBM S2 &  0.3487 &  0.7465 &  0.6433 &    2.0 &        0.0 &     585.15 \\
52  &  NHTS-MW &     14 &        LGBM S1 &  0.3487 &  0.7465 &  0.6433 &   17.0 &        0.0 &    2337.83 \\
53  &  NHTS-MW &     13 &        LGBM S3 &  0.3487 &  0.7465 &  0.6433 &    4.0 &        0.0 &     655.64 \\
54  &  NHTS-MW &     12 &  WV-BATCH &  0.3487 &  0.7465 &  0.6433 &   58.0 &        1.0 &    8219.19 \\
55  &  NHTS-MW &     19 &        LGBM B1 &  0.3487 &  0.7465 &  0.6433 &    0.0 &        0.0 &    3390.72 \\
56  &  NHTS-MW &     20 &          DT B2 &  0.3255 &  0.6127 &  0.4939 &    0.0 &        0.0 &     233.86 \\
57  &  NHTS-MW &     21 &          DT B1 &  0.2952 &  0.6805 &  0.5657 &    0.0 &        0.0 &     329.61 \\
58  &  NHTS-MW &     18 &          DT S2 &  0.2948 &  0.6627 &  0.5416 &    0.0 &        0.0 &     203.15 \\
59  &  NHTS-MW &     22 &          LR B2 &  0.2719 &  0.3981 &  0.2659 &    0.0 &        0.0 &     717.05 \\
60  &  NHTS-MW &     23 &          RF B2 &  0.2320 &  0.5948 &  0.4415 &    0.0 &        0.0 &    2476.18 \\
61  &  NHTS-MW &     27 &            SRP &  0.2097 &  0.7014 &  0.5672 &    0.0 &        0.0 &   70330.61 \\
62  &  NHTS-MW &     25 &          LR B1 &  0.2092 &  0.3503 &  0.2077 &    0.0 &        0.0 &     633.07 \\
63  &  NHTS-MW &     28 &          RF B1 &  0.1886 &  0.6073 &  0.4082 &    0.0 &        0.0 &    3061.13 \\
64  &  NHTS-MW &     24 &        LGBM B2 &  0.1177 &  0.3566 &  0.1949 &    0.0 &        0.0 &    2794.48 \\
65  &  NHTS-MW &     29 &      DS-ONLINE &  0.0901 &  0.3852 &  0.1730 &    0.0 &        0.0 &   83130.95 \\
66  &  NHTS-MW &     31 &            HAT &  0.0835 &  0.3012 &  0.1508 &    0.0 &        0.0 &   36687.71 \\
67  &  NHTS-MW &     26 &      WV-ONLINE &  0.0814 &  0.4320 &  0.1386 &    0.0 &        0.0 &   84483.09 \\
68  &  NHTS-MW &     33 &          NB B2 &  0.0675 &  0.0947 &  0.0208 &    0.0 &        0.0 &    1076.56 \\
69  &  NHTS-MW &     35 &          NB S2 &  0.0611 &  0.1259 &  0.0301 &    3.0 &        0.0 &     918.80 \\
70  &  NHTS-MW &     32 &          NB S3 &  0.0611 &  0.1259 &  0.0301 &    5.0 &        0.0 &    1220.80 \\
71  &  NHTS-MW &     36 &          NB B1 &  0.0611 &  0.1259 &  0.0301 &    0.0 &        0.0 &     681.23 \\
72  &  NHTS-MW &     34 &            ARF &  0.0561 &  0.4690 &  0.1121 &    0.0 &        0.0 &    2560.09 \\
73  &  NHTS-MW &     30 &          NB S1 &  0.0500 &  0.0999 &  0.0281 &   34.0 &       12.0 &    1357.63 \\
74  &  NHTS-MW &     37 &            ONB &  0.0267 &  0.4159 & -0.0000 &    0.0 &        0.0 &   23804.50 \\
75  &  NHTS-MW &     38 &            OLR &  0.0128 &  0.0233 &  0.0107 &    0.0 &        0.0 &    1773.06 \\
\midrule
76  &  NHTS-NE &      1 &          DS-RF &  0.4641 &  0.7792 &  0.6832 &  112.0 &       59.0 &   87765.85 \\
77  &  NHTS-NE &      2 &          WV-RF &  0.4599 &  0.7767 &  0.6777 &  112.0 &       59.0 &   87051.09 \\
78  &  NHTS-NE &      9 &          DT S1 &  0.4357 &  0.7775 &  0.6969 &   38.0 &       17.0 &     307.42 \\
79  &  NHTS-NE &      7 &          RF S1 &  0.4251 &  0.7543 &  0.6476 &   41.0 &       20.0 &    2732.06 \\
80  &  NHTS-NE &     11 &          LR S1 &  0.4207 &  0.6539 &  0.5242 &   37.0 &       18.0 &     669.07 \\
81  &  NHTS-NE &     10 &          DT S3 &  0.4162 &  0.7609 &  0.6738 &   16.0 &        8.0 &     290.92 \\
82  &  NHTS-NE &      4 &          RF S3 &  0.4142 &  0.7555 &  0.6488 &   16.0 &        9.0 &    2528.74 \\
83  &  NHTS-NE &     20 &          DT B2 &  0.4021 &  0.6584 &  0.5486 &    0.0 &        0.0 &     251.94 \\
84  &  NHTS-NE &      5 &  DS-BATCH &  0.3787 &  0.7813 &  0.6948 &   64.0 &        2.0 &    7595.85 \\
85  &  NHTS-NE &      3 &        DS-LGBM &  0.3770 &  0.7851 &  0.6982 &   53.0 &        2.0 &   65082.30 \\
86  &  NHTS-NE &     17 &          LR S2 &  0.3759 &  0.6412 &  0.5009 &   10.0 &        6.0 &     437.09 \\
87  &  NHTS-NE &      8 &          LR S3 &  0.3750 &  0.6415 &  0.5058 &   18.0 &        7.0 &     619.50 \\
88  &  NHTS-NE &      6 &        WV-LGBM &  0.3588 &  0.7791 &  0.6888 &   53.0 &        2.0 &   61123.38 \\
89  &  NHTS-NE &     16 &        LGBM S2 &  0.3580 &  0.7750 &  0.6850 &    1.0 &        0.0 &    1388.38 \\
90  &  NHTS-NE &     14 &        LGBM S1 &  0.3580 &  0.7750 &  0.6850 &   27.0 &        0.0 &    3097.06 \\
91  &  NHTS-NE &     13 &        LGBM S3 &  0.3580 &  0.7750 &  0.6850 &    9.0 &        0.0 &    1126.49 \\
92  &  NHTS-NE &     12 &  WV-BATCH &  0.3580 &  0.7750 &  0.6850 &   64.0 &        2.0 &    6545.09 \\
93  &  NHTS-NE &     19 &        LGBM B1 &  0.3580 &  0.7750 &  0.6850 &    0.0 &        0.0 &    3383.84 \\
94  &  NHTS-NE &     15 &          RF S2 &  0.3252 &  0.7225 &  0.5976 &    4.0 &        4.0 &    2353.08 \\
95  &  NHTS-NE &     18 &          DT S2 &  0.3092 &  0.7174 &  0.6153 &    0.0 &        0.0 &     141.83 \\
96  &  NHTS-NE &     21 &          DT B1 &  0.3089 &  0.7089 &  0.6031 &    0.0 &        0.0 &     345.22 \\
97  &  NHTS-NE &     23 &          RF B2 &  0.2787 &  0.6444 &  0.5067 &    0.0 &        0.0 &    2502.35 \\
98  &  NHTS-NE &     22 &          LR B2 &  0.2683 &  0.4299 &  0.2942 &    0.0 &        0.0 &     808.16 \\
99  &  NHTS-NE &     28 &          RF B1 &  0.2528 &  0.6730 &  0.5200 &    0.0 &        0.0 &    3169.13 \\
100 &  NHTS-NE &     27 &            SRP &  0.2313 &  0.7322 &  0.6144 &    0.0 &        0.0 &   55738.82 \\
101 &  NHTS-NE &     25 &          LR B1 &  0.2028 &  0.4620 &  0.3152 &    0.0 &        0.0 &     714.31 \\
102 &  NHTS-NE &     24 &        LGBM B2 &  0.1666 &  0.5065 &  0.3584 &    0.0 &        0.0 &    3561.00 \\
103 &  NHTS-NE &     29 &      DS-ONLINE &  0.0845 &  0.4079 &  0.1869 &    0.0 &        0.0 &   78882.46 \\
104 &  NHTS-NE &     31 &            HAT &  0.0829 &  0.3189 &  0.1707 &    0.0 &        0.0 &   30021.90 \\
105 &  NHTS-NE &     26 &      WV-ONLINE &  0.0772 &  0.4524 &  0.1551 &    0.0 &        0.0 &   76859.47 \\
106 &  NHTS-NE &     33 &          NB B2 &  0.0668 &  0.0839 &  0.0149 &    0.0 &        0.0 &     960.97 \\
107 &  NHTS-NE &     32 &          NB S3 &  0.0655 &  0.0867 &  0.0261 &   15.0 &        5.0 &     513.10 \\
108 &  NHTS-NE &     34 &            ARF &  0.0614 &  0.4816 &  0.1269 &    0.0 &        0.0 &    1463.03 \\
109 &  NHTS-NE &     30 &          NB S1 &  0.0589 &  0.0866 &  0.0285 &   43.0 &       14.0 &     523.56 \\
110 &  NHTS-NE &     35 &          NB S2 &  0.0511 &  0.0811 &  0.0247 &    5.0 &        0.0 &    1560.00 \\
111 &  NHTS-NE &     36 &          NB B1 &  0.0511 &  0.0811 &  0.0247 &    0.0 &        0.0 &     788.83 \\
112 &  NHTS-NE &     37 &            ONB &  0.0272 &  0.4279 & -0.0000 &    0.0 &        0.0 &   23059.12 \\
113 &  NHTS-NE &     38 &            OLR &  0.0117 &  0.0250 &  0.0135 &    0.0 &        0.0 &    1619.25 \\
\midrule
114 &  NHTS-SE &      2 &          WV-RF &  0.4557 &  0.7653 &  0.6543 &  168.0 &       85.0 &  118115.13 \\
115 &  NHTS-SE &      1 &          DS-RF &  0.4530 &  0.7662 &  0.6586 &  168.0 &       85.0 &  118014.45 \\
116 &  NHTS-SE &     15 &          RF S2 &  0.4285 &  0.7432 &  0.6233 &   18.0 &       16.0 &    3600.94 \\
117 &  NHTS-SE &      4 &          RF S3 &  0.4285 &  0.7432 &  0.6233 &   23.0 &       16.0 &    3690.82 \\
118 &  NHTS-SE &      9 &          DT S1 &  0.4277 &  0.7641 &  0.6738 &   47.0 &       24.0 &     525.47 \\
119 &  NHTS-SE &     18 &          DT S2 &  0.4271 &  0.7564 &  0.6628 &   16.0 &       12.0 &     366.81 \\
120 &  NHTS-SE &     17 &          LR S2 &  0.4208 &  0.6426 &  0.4979 &   12.0 &       10.0 &     580.04 \\
121 &  NHTS-SE &      8 &          LR S3 &  0.4201 &  0.6400 &  0.4972 &   12.0 &       10.0 &     573.24 \\
122 &  NHTS-SE &      7 &          RF S1 &  0.4125 &  0.7389 &  0.6166 &   61.0 &       28.0 &    4028.50 \\
123 &  NHTS-SE &     10 &          DT S3 &  0.4061 &  0.7446 &  0.6465 &   12.0 &        7.0 &     427.95 \\
124 &  NHTS-SE &     11 &          LR S1 &  0.4012 &  0.6290 &  0.4866 &   55.0 &       28.0 &     998.26 \\
125 &  NHTS-SE &      6 &        WV-LGBM &  0.3165 &  0.7544 &  0.6478 &   91.0 &        2.0 &  108938.05 \\
126 &  NHTS-SE &      3 &        DS-LGBM &  0.3159 &  0.7522 &  0.6455 &   91.0 &        2.0 &  119122.05 \\
127 &  NHTS-SE &     20 &          DT B2 &  0.3151 &  0.6398 &  0.5197 &    0.0 &        0.0 &     591.60 \\
128 &  NHTS-SE &     16 &        LGBM S2 &  0.3141 &  0.7509 &  0.6446 &    2.0 &        0.0 &     476.56 \\
129 &  NHTS-SE &     13 &        LGBM S3 &  0.3141 &  0.7509 &  0.6446 &    7.0 &        0.0 &     996.73 \\
130 &  NHTS-SE &     12 &  WV-BATCH &  0.3141 &  0.7509 &  0.6446 &   91.0 &        2.0 &   11946.93 \\
131 &  NHTS-SE &     19 &        LGBM B1 &  0.3141 &  0.7509 &  0.6446 &    0.0 &        0.0 &    4694.60 \\
132 &  NHTS-SE &      5 &  DS-BATCH &  0.3135 &  0.7501 &  0.6436 &   91.0 &        2.0 &   11022.63 \\
133 &  NHTS-SE &     22 &          LR B2 &  0.2646 &  0.4001 &  0.2605 &    0.0 &        0.0 &    1315.21 \\
134 &  NHTS-SE &     21 &          DT B1 &  0.2400 &  0.6682 &  0.5370 &    0.0 &        0.0 &     315.66 \\
135 &  NHTS-SE &     23 &          RF B2 &  0.2200 &  0.6321 &  0.4746 &    0.0 &        0.0 &    4485.57 \\
136 &  NHTS-SE &     25 &          LR B1 &  0.2093 &  0.3930 &  0.2317 &    0.0 &        0.0 &     279.89 \\
137 &  NHTS-SE &     14 &        LGBM S1 &  0.2049 &  0.6460 &  0.5059 &   36.0 &        6.0 &    1240.84 \\
138 &  NHTS-SE &     27 &            SRP &  0.1947 &  0.6935 &  0.5481 &    0.0 &        0.0 &   92934.53 \\
139 &  NHTS-SE &     28 &          RF B1 &  0.1670 &  0.6251 &  0.4237 &    0.0 &        0.0 &    4228.90 \\
140 &  NHTS-SE &     24 &        LGBM B2 &  0.1424 &  0.4900 &  0.3432 &    0.0 &        0.0 &    4700.93 \\
141 &  NHTS-SE &     29 &      DS-ONLINE &  0.0775 &  0.3746 &  0.1400 &    0.0 &        0.0 &  124328.69 \\
142 &  NHTS-SE &     31 &            HAT &  0.0691 &  0.2547 &  0.1085 &    0.0 &        0.0 &   51264.47 \\
143 &  NHTS-SE &     26 &      WV-ONLINE &  0.0686 &  0.4416 &  0.1202 &    0.0 &        0.0 &  126428.51 \\
144 &  NHTS-SE &     33 &          NB B2 &  0.0675 &  0.0800 &  0.0151 &    0.0 &        0.0 &    1280.62 \\
145 &  NHTS-SE &     30 &          NB S1 &  0.0603 &  0.0885 &  0.0234 &   63.0 &       24.0 &     938.44 \\
146 &  NHTS-SE &     34 &            ARF &  0.0514 &  0.4801 &  0.1063 &    0.0 &        0.0 &    2179.29 \\
147 &  NHTS-SE &     32 &          NB S3 &  0.0447 &  0.0763 &  0.0214 &   14.0 &        7.0 &     802.59 \\
148 &  NHTS-SE &     35 &          NB S2 &  0.0433 &  0.0745 &  0.0197 &    9.0 &        5.0 &     693.19 \\
149 &  NHTS-SE &     36 &          NB B1 &  0.0398 &  0.1059 &  0.0226 &    0.0 &        0.0 &     686.33 \\
150 &  NHTS-SE &     37 &            ONB &  0.0274 &  0.4325 & -0.0000 &    0.0 &        0.0 &   36869.95 \\
151 &  NHTS-SE &     38 &            OLR &  0.0108 &  0.0154 &  0.0083 &    0.0 &        0.0 &    2448.38 \\
\midrule
152 &  NHTS-SW &      1 &          DS-RF &  0.4643 &  0.7807 &  0.6864 &  151.0 &       80.0 &   99704.80 \\
153 &  NHTS-SW &      2 &          WV-RF &  0.4595 &  0.7790 &  0.6819 &  151.0 &       80.0 &  102847.74 \\
154 &  NHTS-SW &      4 &          RF S3 &  0.4211 &  0.7587 &  0.6553 &   21.0 &       14.0 &    3345.15 \\
155 &  NHTS-SW &      8 &          LR S3 &  0.4048 &  0.6290 &  0.4918 &   18.0 &       14.0 &     608.19 \\
156 &  NHTS-SW &     11 &          LR S1 &  0.4021 &  0.6386 &  0.5018 &   47.0 &       22.0 &    1398.39 \\
157 &  NHTS-SW &      9 &          DT S1 &  0.3939 &  0.7632 &  0.6747 &   54.0 &       25.0 &     433.37 \\
158 &  NHTS-SW &      7 &          RF S1 &  0.3920 &  0.7454 &  0.6343 &   54.0 &       26.0 &    3677.16 \\
159 &  NHTS-SW &     10 &          DT S3 &  0.3859 &  0.7514 &  0.6591 &   13.0 &        8.0 &     310.39 \\
160 &  NHTS-SW &     18 &          DT S2 &  0.3818 &  0.7484 &  0.6551 &    8.0 &        7.0 &     242.74 \\
161 &  NHTS-SW &      3 &        DS-LGBM &  0.3584 &  0.7827 &  0.6970 &   93.0 &       19.0 &   85107.34 \\
162 &  NHTS-SW &      5 &  DS-BATCH &  0.3574 &  0.7789 &  0.6934 &   93.0 &       19.0 &    7613.78 \\
163 &  NHTS-SW &      6 &        WV-LGBM &  0.3568 &  0.7873 &  0.7012 &   93.0 &       19.0 &  104524.58 \\
164 &  NHTS-SW &     12 &  WV-BATCH &  0.3557 &  0.7833 &  0.6984 &   93.0 &       19.0 &    7228.98 \\
165 &  NHTS-SW &     20 &          DT B2 &  0.3469 &  0.6540 &  0.5398 &    0.0 &        0.0 &     249.26 \\
166 &  NHTS-SW &     17 &          LR S2 &  0.3442 &  0.6132 &  0.4684 &    9.0 &        7.0 &     551.88 \\
167 &  NHTS-SW &     16 &        LGBM S2 &  0.3391 &  0.7721 &  0.6808 &    1.0 &        0.0 &     596.61 \\
168 &  NHTS-SW &     13 &        LGBM S3 &  0.3391 &  0.7721 &  0.6808 &    6.0 &        0.0 &     928.74 \\
169 &  NHTS-SW &     19 &        LGBM B1 &  0.3391 &  0.7721 &  0.6808 &    0.0 &        0.0 &    4628.27 \\
170 &  NHTS-SW &     14 &        LGBM S1 &  0.3364 &  0.7710 &  0.6793 &   38.0 &        1.0 &    5438.19 \\
171 &  NHTS-SW &     15 &          RF S2 &  0.2992 &  0.7069 &  0.5762 &    7.0 &        5.0 &    3343.71 \\
172 &  NHTS-SW &     22 &          LR B2 &  0.2783 &  0.5194 &  0.3395 &    0.0 &        0.0 &    1243.45 \\
173 &  NHTS-SW &     21 &          DT B1 &  0.2543 &  0.6982 &  0.5878 &    0.0 &        0.0 &     524.11 \\
174 &  NHTS-SW &     23 &          RF B2 &  0.2289 &  0.6303 &  0.4847 &    0.0 &        0.0 &    4704.03 \\
175 &  NHTS-SW &     28 &          RF B1 &  0.2110 &  0.6742 &  0.5250 &    0.0 &        0.0 &    4091.52 \\
176 &  NHTS-SW &     25 &          LR B1 &  0.2108 &  0.4252 &  0.2500 &    0.0 &        0.0 &     814.19 \\
177 &  NHTS-SW &     24 &        LGBM B2 &  0.1787 &  0.5422 &  0.3811 &    0.0 &        0.0 &    5410.17 \\
178 &  NHTS-SW &     27 &            SRP &  0.1613 &  0.6864 &  0.5467 &    0.0 &        0.0 &   79355.50 \\
179 &  NHTS-SW &     29 &      DS-ONLINE &  0.0803 &  0.4301 &  0.1710 &    0.0 &        0.0 &  109513.14 \\
180 &  NHTS-SW &     26 &      WV-ONLINE &  0.0711 &  0.4591 &  0.1493 &    0.0 &        0.0 &  109645.59 \\
181 &  NHTS-SW &     31 &            HAT &  0.0637 &  0.2021 &  0.0831 &    0.0 &        0.0 &   57760.73 \\
182 &  NHTS-SW &     34 &            ARF &  0.0611 &  0.4957 &  0.1767 &    0.0 &        0.0 &    2899.39 \\
183 &  NHTS-SW &     35 &          NB S2 &  0.0565 &  0.0819 &  0.0279 &    9.0 &        4.0 &     727.22 \\
184 &  NHTS-SW &     30 &          NB S1 &  0.0554 &  0.0750 &  0.0270 &   58.0 &       25.0 &     691.29 \\
185 &  NHTS-SW &     32 &          NB S3 &  0.0534 &  0.0674 &  0.0267 &   17.0 &        9.0 &     639.35 \\
186 &  NHTS-SW &     36 &          NB B1 &  0.0506 &  0.0837 &  0.0285 &    0.0 &        0.0 &     937.72 \\
187 &  NHTS-SW &     33 &          NB B2 &  0.0470 &  0.0571 &  0.0185 &    0.0 &        0.0 &     704.75 \\
188 &  NHTS-SW &     37 &            ONB &  0.0263 &  0.4074 & -0.0000 &    0.0 &        0.0 &   38147.17 \\
189 &  NHTS-SW &     38 &            OLR &  0.0114 &  0.0152 &  0.0083 &    0.0 &        0.0 &    3322.76 \\
\midrule
190 &   NHTS-W &     10 &          DT S3 &  0.4848 &  0.7722 &  0.6836 &   28.0 &       14.0 &     566.05 \\
191 &   NHTS-W &      1 &          DS-RF &  0.4532 &  0.7744 &  0.6654 &  187.0 &       96.0 &  143264.35 \\
192 &   NHTS-W &      9 &          DT S1 &  0.4527 &  0.7682 &  0.6780 &   50.0 &       25.0 &     594.09 \\
193 &   NHTS-W &      2 &          WV-RF &  0.4353 &  0.7687 &  0.6537 &  187.0 &       96.0 &  141365.07 \\
194 &   NHTS-W &     15 &          RF S2 &  0.4173 &  0.7470 &  0.6217 &   20.0 &       17.0 &    4178.44 \\
195 &   NHTS-W &      4 &          RF S3 &  0.4173 &  0.7470 &  0.6217 &   30.0 &       17.0 &    4214.24 \\
196 &   NHTS-W &     11 &          LR S1 &  0.4115 &  0.6556 &  0.5178 &   47.0 &       29.0 &     897.65 \\
197 &   NHTS-W &      8 &          LR S3 &  0.4095 &  0.6627 &  0.5189 &   21.0 &       12.0 &     902.13 \\
198 &   NHTS-W &      7 &          RF S1 &  0.3808 &  0.7412 &  0.6121 &   66.0 &       34.0 &    4245.50 \\
199 &   NHTS-W &     17 &          LR S2 &  0.3802 &  0.6363 &  0.4906 &    9.0 &        6.0 &     628.01 \\
200 &   NHTS-W &     20 &          DT B2 &  0.3685 &  0.6631 &  0.5399 &    0.0 &        0.0 &     421.93 \\
201 &   NHTS-W &     18 &          DT S2 &  0.3432 &  0.6910 &  0.5743 &    0.0 &        0.0 &     198.67 \\
202 &   NHTS-W &     21 &          DT B1 &  0.3390 &  0.6877 &  0.5709 &    0.0 &        0.0 &     559.44 \\
203 &   NHTS-W &      5 &  DS-BATCH &  0.3204 &  0.7635 &  0.6590 &   99.0 &        8.0 &    7590.45 \\
204 &   NHTS-W &      3 &        DS-LGBM &  0.3200 &  0.7661 &  0.6615 &   99.0 &        8.0 &  123308.00 \\
205 &   NHTS-W &     14 &        LGBM S1 &  0.3175 &  0.7481 &  0.6406 &   36.0 &        4.0 &    2099.50 \\
206 &   NHTS-W &     12 &  WV-BATCH &  0.3072 &  0.7589 &  0.6516 &   99.0 &        8.0 &    8258.08 \\
207 &   NHTS-W &      6 &        WV-LGBM &  0.3064 &  0.7638 &  0.6556 &   99.0 &        8.0 &  110823.18 \\
208 &   NHTS-W &     16 &        LGBM S2 &  0.3024 &  0.7546 &  0.6454 &    3.0 &        0.0 &     976.63 \\
209 &   NHTS-W &     13 &        LGBM S3 &  0.3024 &  0.7546 &  0.6454 &   13.0 &        0.0 &    2646.26 \\
210 &   NHTS-W &     19 &        LGBM B1 &  0.3024 &  0.7546 &  0.6454 &    0.0 &        0.0 &    5477.02 \\
211 &   NHTS-W &     22 &          LR B2 &  0.2538 &  0.4007 &  0.2539 &    0.0 &        0.0 &     953.67 \\
212 &   NHTS-W &     27 &            SRP &  0.2096 &  0.7146 &  0.5687 &    0.0 &        0.0 &   96165.77 \\
213 &   NHTS-W &     23 &          RF B2 &  0.1970 &  0.6438 &  0.4712 &    0.0 &        0.0 &    4565.17 \\
214 &   NHTS-W &     25 &          LR B1 &  0.1921 &  0.4242 &  0.2373 &    0.0 &        0.0 &    1291.27 \\
215 &   NHTS-W &     24 &        LGBM B2 &  0.1831 &  0.5814 &  0.4264 &    0.0 &        0.0 &    4757.49 \\
216 &   NHTS-W &     28 &          RF B1 &  0.1465 &  0.6215 &  0.3937 &    0.0 &        0.0 &    5196.77 \\
217 &   NHTS-W &     29 &      DS-ONLINE &  0.0739 &  0.3853 &  0.1586 &    0.0 &        0.0 &  133554.11 \\
218 &   NHTS-W &     33 &          NB B2 &  0.0723 &  0.0575 &  0.0176 &    0.0 &        0.0 &     885.35 \\
219 &   NHTS-W &     31 &            HAT &  0.0717 &  0.2745 &  0.1379 &    0.0 &        0.0 &   52504.42 \\
220 &   NHTS-W &     26 &      WV-ONLINE &  0.0656 &  0.4623 &  0.1293 &    0.0 &        0.0 &  131808.03 \\
221 &   NHTS-W &     30 &          NB S1 &  0.0605 &  0.0861 &  0.0297 &   64.0 &       17.0 &    1177.87 \\
222 &   NHTS-W &     32 &          NB S3 &  0.0554 &  0.0774 &  0.0256 &   25.0 &       11.0 &     966.08 \\
223 &   NHTS-W &     35 &          NB S2 &  0.0552 &  0.0815 &  0.0266 &   13.0 &        7.0 &     887.78 \\
224 &   NHTS-W &     36 &          NB B1 &  0.0465 &  0.0752 &  0.0194 &    0.0 &        0.0 &     662.44 \\
225 &   NHTS-W &     34 &            ARF &  0.0455 &  0.4828 &  0.0712 &    0.0 &        0.0 &    2759.94 \\
226 &   NHTS-W &     37 &            ONB &  0.0284 &  0.4553 &  0.0000 &    0.0 &        0.0 &   33103.23 \\
227 &   NHTS-W &     38 &            OLR &  0.0135 &  0.0327 &  0.0150 &    0.0 &        0.0 &    2684.69 \\
\midrule
228 &      NTS &      6 &        WV-LGBM &  0.5415 &  0.6787 &  0.4374 &  141.0 &       39.0 &    2601.83 \\
229 &      NTS &      5 &  DS-BATCH &  0.5412 &  0.6642 &  0.4229 &   57.0 &       24.0 &    1314.61 \\
230 &      NTS &      3 &        DS-LGBM &  0.5397 &  0.6655 &  0.4257 &  141.0 &       39.0 &    2706.23 \\
231 &      NTS &     12 &  WV-BATCH &  0.5386 &  0.6658 &  0.4241 &   57.0 &       24.0 &    1192.17 \\
232 &      NTS &      2 &          WV-RF &  0.5345 &  0.6782 &  0.4312 &  183.0 &       57.0 &   16451.08 \\
233 &      NTS &     14 &        LGBM S1 &  0.5343 &  0.6602 &  0.4141 &   54.0 &       16.0 &     354.19 \\
234 &      NTS &      1 &          DS-RF &  0.5324 &  0.6673 &  0.4226 &  183.0 &       57.0 &   16392.98 \\
235 &      NTS &     13 &        LGBM S3 &  0.5299 &  0.6571 &  0.4087 &   25.0 &       11.0 &     313.80 \\
236 &      NTS &      7 &          RF S1 &  0.5260 &  0.6628 &  0.4108 &   53.0 &       15.0 &    4568.37 \\
237 &      NTS &      4 &          RF S3 &  0.5202 &  0.6587 &  0.4056 &   25.0 &        7.0 &    3900.09 \\
238 &      NTS &     16 &        LGBM S2 &  0.5107 &  0.6465 &  0.3880 &    2.0 &        2.0 &     281.23 \\
239 &      NTS &     24 &        LGBM B2 &  0.5059 &  0.6253 &  0.3754 &    0.0 &        0.0 &    3280.94 \\
240 &      NTS &     15 &          RF S2 &  0.4916 &  0.6415 &  0.3790 &    2.0 &        2.0 &    4200.99 \\
241 &      NTS &     23 &          RF B2 &  0.4855 &  0.6218 &  0.3626 &    0.0 &        0.0 &    4014.39 \\
242 &      NTS &      8 &          LR S3 &  0.4813 &  0.6437 &  0.3724 &   25.0 &       12.0 &     225.38 \\
243 &      NTS &     26 &      WV-ONLINE &  0.4806 &  0.6393 &  0.3673 &    0.0 &        0.0 &    1184.56 \\
244 &      NTS &     11 &          LR S1 &  0.4668 &  0.6357 &  0.3539 &   53.0 &       24.0 &     267.63 \\
245 &      NTS &     17 &          LR S2 &  0.4547 &  0.6362 &  0.3639 &    2.0 &        1.0 &     166.85 \\
246 &      NTS &     22 &          LR B2 &  0.4403 &  0.6026 &  0.3220 &    0.0 &        0.0 &     143.17 \\
247 &      NTS &     32 &          NB S3 &  0.4280 &  0.5330 &  0.2650 &   24.0 &        9.0 &     213.61 \\
248 &      NTS &     10 &          DT S3 &  0.4236 &  0.5380 &  0.2513 &   24.0 &       11.0 &     167.05 \\
249 &      NTS &      9 &          DT S1 &  0.4235 &  0.5360 &  0.2513 &   53.0 &       23.0 &     207.70 \\
250 &      NTS &     30 &          NB S1 &  0.4233 &  0.5268 &  0.2572 &   52.0 &       15.0 &     270.74 \\
251 &      NTS &     20 &          DT B2 &  0.4104 &  0.5138 &  0.2319 &    0.0 &        0.0 &     105.36 \\
252 &      NTS &     33 &          NB B2 &  0.4072 &  0.5040 &  0.2351 &    0.0 &        0.0 &     155.73 \\
253 &      NTS &     29 &      DS-ONLINE &  0.3751 &  0.6511 &  0.3742 &    0.0 &        0.0 &    1313.52 \\
254 &      NTS &     27 &            SRP &  0.3748 &  0.6486 &  0.3625 &    0.0 &        0.0 &    1091.14 \\
255 &      NTS &     34 &            ARF &  0.3727 &  0.6766 &  0.3961 &    0.0 &        0.0 &    1152.61 \\
256 &      NTS &     35 &          NB S2 &  0.3643 &  0.4481 &  0.2106 &    0.0 &        0.0 &     176.59 \\
257 &      NTS &     36 &          NB B1 &  0.3643 &  0.4481 &  0.2106 &    0.0 &        0.0 &     147.09 \\
258 &      NTS &     25 &          LR B1 &  0.3633 &  0.6037 &  0.2988 &    0.0 &        0.0 &     132.41 \\
259 &      NTS &     28 &          RF B1 &  0.3555 &  0.5957 &  0.2969 &    0.0 &        0.0 &    4155.82 \\
260 &      NTS &     19 &        LGBM B1 &  0.3552 &  0.5817 &  0.2852 &    0.0 &        0.0 &    4552.39 \\
261 &      NTS &     21 &          DT B1 &  0.3492 &  0.4916 &  0.1784 &    0.0 &        0.0 &     141.69 \\
262 &      NTS &     18 &          DT S2 &  0.3488 &  0.4899 &  0.1753 &    0.0 &        0.0 &     123.73 \\
263 &      NTS &     37 &            ONB &  0.3477 &  0.5436 &  0.2793 &    0.0 &        0.0 &     153.35 \\
264 &      NTS &     31 &            HAT &  0.2724 &  0.4438 &  0.1682 &    0.0 &        0.0 &     473.12 \\
265 &      NTS &     38 &            OLR &  0.1799 &  0.5534 &  0.0026 &    0.0 &        0.0 &      80.24 \\
\midrule
266 &     Ohio &      5 &  DS-BATCH &  0.2242 &  0.8724 &  0.6808 &   50.0 &        8.0 &    1074.72 \\
267 &     Ohio &      3 &        DS-LGBM &  0.2239 &  0.8728 &  0.6824 &   94.0 &        9.0 &    7419.69 \\
268 &     Ohio &     14 &        LGBM S1 &  0.2173 &  0.8674 &  0.6678 &   34.0 &        2.0 &    1732.27 \\
269 &     Ohio &     10 &          DT S3 &  0.2164 &  0.8233 &  0.5893 &   13.0 &        9.0 &    1722.45 \\
270 &     Ohio &     12 &  WV-BATCH &  0.2156 &  0.8686 &  0.6684 &   50.0 &        8.0 &    1004.59 \\
271 &     Ohio &      6 &        WV-LGBM &  0.2146 &  0.8707 &  0.6685 &   94.0 &        9.0 &    7295.75 \\
272 &     Ohio &      9 &          DT S1 &  0.2138 &  0.8192 &  0.5756 &   34.0 &       17.0 &    1782.37 \\
273 &     Ohio &     16 &        LGBM S2 &  0.2130 &  0.8664 &  0.6623 &    4.0 &        0.0 &     193.32 \\
274 &     Ohio &     13 &        LGBM S3 &  0.2130 &  0.8664 &  0.6623 &   15.0 &        0.0 &    1501.06 \\
275 &     Ohio &     19 &        LGBM B1 &  0.2130 &  0.8664 &  0.6623 &    0.0 &        0.0 &    2354.07 \\
276 &     Ohio &      8 &          LR S3 &  0.2068 &  0.8412 &  0.6110 &   12.0 &        7.0 &    1726.50 \\
277 &     Ohio &      1 &          DS-RF &  0.2059 &  0.8712 &  0.6601 &  124.0 &       52.0 &   16161.06 \\
278 &     Ohio &     18 &          DT S2 &  0.2051 &  0.8179 &  0.5756 &    3.0 &        2.0 &     111.39 \\
279 &     Ohio &      4 &          RF S3 &  0.2051 &  0.8706 &  0.6581 &   16.0 &        7.0 &    4002.08 \\
280 &     Ohio &     11 &          LR S1 &  0.2021 &  0.8376 &  0.6024 &   36.0 &       17.0 &    1992.12 \\
281 &     Ohio &     22 &          LR B2 &  0.2014 &  0.8155 &  0.5206 &    0.0 &        0.0 &      99.06 \\
282 &     Ohio &      7 &          RF S1 &  0.2002 &  0.8659 &  0.6421 &   38.0 &       18.0 &    4093.52 \\
283 &     Ohio &     21 &          DT B1 &  0.1975 &  0.8189 &  0.5752 &    0.0 &        0.0 &      81.40 \\
284 &     Ohio &      2 &          WV-RF &  0.1970 &  0.8658 &  0.6390 &  124.0 &       52.0 &   15978.95 \\
285 &     Ohio &     23 &          RF B2 &  0.1961 &  0.8478 &  0.5767 &    0.0 &        0.0 &    1841.63 \\
286 &     Ohio &     17 &          LR S2 &  0.1877 &  0.8159 &  0.5585 &    3.0 &        2.0 &     153.50 \\
287 &     Ohio &     20 &          DT B2 &  0.1873 &  0.7963 &  0.5039 &    0.0 &        0.0 &      73.81 \\
288 &     Ohio &     15 &          RF S2 &  0.1871 &  0.8600 &  0.6269 &    5.0 &        5.0 &    2372.49 \\
289 &     Ohio &     27 &            SRP &  0.1771 &  0.8658 &  0.6481 &    0.0 &        0.0 &    4149.69 \\
290 &     Ohio &     25 &          LR B1 &  0.1756 &  0.7879 &  0.4946 &    0.0 &        0.0 &     104.80 \\
291 &     Ohio &     26 &      WV-ONLINE &  0.1660 &  0.8439 &  0.5689 &    0.0 &        0.0 &    2411.53 \\
292 &     Ohio &     28 &          RF B1 &  0.1636 &  0.8503 &  0.5969 &    0.0 &        0.0 &    2262.79 \\
293 &     Ohio &     29 &      DS-ONLINE &  0.1619 &  0.8465 &  0.5773 &    0.0 &        0.0 &    2506.60 \\
294 &     Ohio &     24 &        LGBM B2 &  0.1616 &  0.7557 &  0.4283 &    0.0 &        0.0 &    1925.58 \\
295 &     Ohio &     34 &            ARF &  0.1573 &  0.8483 &  0.5806 &    0.0 &        0.0 &     722.21 \\
296 &     Ohio &     38 &            OLR &  0.1217 &  0.7947 &  0.4823 &    0.0 &        0.0 &     176.67 \\
297 &     Ohio &     37 &            ONB &  0.1131 &  0.8156 &  0.4839 &    0.0 &        0.0 &    1170.22 \\
298 &     Ohio &     31 &            HAT &  0.0780 &  0.3376 &  0.1058 &    0.0 &        0.0 &    1858.16 \\
299 &     Ohio &     33 &          NB B2 &  0.0711 &  0.3344 &  0.0759 &    0.0 &        0.0 &     141.96 \\
300 &     Ohio &     35 &          NB S2 &  0.0629 &  0.2798 &  0.0848 &    7.0 &        2.0 &     200.55 \\
301 &     Ohio &     32 &          NB S3 &  0.0629 &  0.2798 &  0.0848 &   18.0 &        2.0 &    1951.37 \\
302 &     Ohio &     30 &          NB S1 &  0.0629 &  0.2798 &  0.0848 &   42.0 &        2.0 &    2091.38 \\
303 &     Ohio &     36 &          NB B1 &  0.0598 &  0.3172 &  0.0886 &    0.0 &        0.0 &     161.69 \\
\midrule
304 &   Optima &     12 &  WV-BATCH &  0.4590 &  0.6110 &  0.3013 &  117.0 &       22.0 &    2541.97 \\
305 &   Optima &     13 &        LGBM S3 &  0.4533 &  0.5894 &  0.2905 &   17.0 &        4.0 &     717.30 \\
306 &   Optima &      5 &  DS-BATCH &  0.4461 &  0.5700 &  0.2606 &  117.0 &       22.0 &    2548.02 \\
307 &   Optima &      6 &        WV-LGBM &  0.4384 &  0.6547 &  0.3379 &  117.0 &       22.0 &    8179.01 \\
308 &   Optima &      3 &        DS-LGBM &  0.4365 &  0.6013 &  0.2922 &  117.0 &       22.0 &    2684.22 \\
309 &   Optima &     16 &        LGBM S2 &  0.4286 &  0.5603 &  0.2460 &   51.0 &        8.0 &     926.28 \\
310 &   Optima &     14 &        LGBM S1 &  0.4283 &  0.5603 &  0.2460 &   17.0 &        7.0 &      12.04 \\
311 &   Optima &     18 &          DT S2 &  0.3943 &  0.5082 &  0.2047 &   47.0 &       10.0 &     978.93 \\
312 &   Optima &      1 &          DS-RF &  0.3933 &  0.5978 &  0.2797 &  112.0 &       29.0 &    2997.41 \\
313 &   Optima &     15 &          RF S2 &  0.3915 &  0.5974 &  0.2547 &   49.0 &        8.0 &    1019.55 \\
314 &   Optima &      7 &          RF S1 &  0.3835 &  0.5947 &  0.2527 &   13.0 &        9.0 &      45.91 \\
315 &   Optima &      4 &          RF S3 &  0.3823 &  0.5907 &  0.2264 &   15.0 &        4.0 &     784.60 \\
316 &   Optima &      8 &          LR S3 &  0.3724 &  0.5395 &  0.2296 &   18.0 &        5.0 &     756.07 \\
317 &   Optima &      2 &          WV-RF &  0.3617 &  0.6340 &  0.2725 &  112.0 &       29.0 &    3000.15 \\
318 &   Optima &     10 &          DT S3 &  0.3539 &  0.4627 &  0.1647 &   18.0 &        3.0 &     751.16 \\
319 &   Optima &      9 &          DT S1 &  0.3537 &  0.4693 &  0.1478 &   13.0 &        7.0 &       9.89 \\
320 &   Optima &     11 &          LR S1 &  0.3413 &  0.5139 &  0.1734 &   14.0 &       10.0 &      13.09 \\
321 &   Optima &     26 &      WV-ONLINE &  0.3384 &  0.6137 &  0.2360 &    0.0 &        0.0 &      90.86 \\
322 &   Optima &     17 &          LR S2 &  0.3293 &  0.5161 &  0.1579 &   47.0 &       16.0 &     974.94 \\
323 &   Optima &     30 &          NB S1 &  0.3139 &  0.4830 &  0.1445 &   17.0 &        7.0 &      10.76 \\
324 &   Optima &     29 &      DS-ONLINE &  0.3115 &  0.5700 &  0.2972 &    0.0 &        0.0 &      70.54 \\
325 &   Optima &     19 &        LGBM B1 &  0.3070 &  0.5143 &  0.0843 &    0.0 &        0.0 &      34.98 \\
326 &   Optima &     31 &            HAT &  0.3057 &  0.5492 &  0.2784 &    0.0 &        0.0 &      43.67 \\
327 &   Optima &     21 &          DT B1 &  0.3048 &  0.4344 &  0.0974 &    0.0 &        0.0 &       2.67 \\
328 &   Optima &     35 &          NB S2 &  0.3030 &  0.4728 &  0.1141 &   50.0 &       11.0 &     974.51 \\
329 &   Optima &     22 &          LR B2 &  0.3002 &  0.4552 &  0.1467 &    0.0 &        0.0 &       2.15 \\
330 &   Optima &     24 &        LGBM B2 &  0.2989 &  0.5042 &  0.1263 &    0.0 &        0.0 &      31.50 \\
331 &   Optima &     32 &          NB S3 &  0.2929 &  0.4464 &  0.0892 &   14.0 &        3.0 &     750.43 \\
332 &   Optima &     25 &          LR B1 &  0.2911 &  0.4949 &  0.1185 &    0.0 &        0.0 &       2.86 \\
333 &   Optima &     20 &          DT B2 &  0.2813 &  0.4026 &  0.0965 &    0.0 &        0.0 &       2.17 \\
334 &   Optima &     23 &          RF B2 &  0.2739 &  0.5064 &  0.0941 &    0.0 &        0.0 &      29.85 \\
335 &   Optima &     37 &            ONB &  0.2643 &  0.6137 &  0.2252 &    0.0 &        0.0 &      20.02 \\
336 &   Optima &     33 &          NB B2 &  0.2622 &  0.4358 &  0.0686 &    0.0 &        0.0 &       2.10 \\
337 &   Optima &     27 &            SRP &  0.2340 &  0.5907 &  0.1727 &    0.0 &        0.0 &     135.40 \\
338 &   Optima &     36 &          NB B1 &  0.2161 &  0.4680 & -0.0100 &    0.0 &        0.0 &       2.54 \\
339 &   Optima &     28 &          RF B1 &  0.1991 &  0.5143 & -0.0218 &    0.0 &        0.0 &      33.54 \\
340 &   Optima &     34 &            ARF &  0.1883 &  0.5691 &  0.0677 &    0.0 &        0.0 &      15.12 \\
341 &   Optima &     38 &            OLR &  0.1495 &  0.2512 &  0.0251 &    0.0 &        0.0 &       7.01 \\
\bottomrule

\end{longtable}
\pagebreak